# PyDCM: Custom Data Center Models with Reinforcement Learning for Sustainability

Avisek Naug*
Antonio Guillen*
Ricardo Luna Gutiérrez*
Vineet Gundecha*
avisek.naug@hpe.com
antonio.guillen@hpe.com
rluna@hpe.com
vineet.gundecha@hpe.com
Hewlett Packard Enterprise
San Jose, USA

Dejan Markovikj
Lekhapriya Dheeraj Kashyap
Lorenz Krause
dejan.markovikj@hpe.com
lekhapriya.dheeraj-kashyap@hpe.com
lorenz.krause@hpe.com
Hewlett Packard Enterprise
San Jose, USA

Sahand Ghorbanpour
Sajad Mousavi
Ashwin Ramesh Babu
Soumyendu Sarkar*†
sahand.ghorbanpour@hpe.com
sajad.mousavi@hpe.com
ashwin.ramesh-babu@hpe.com
soumyendu.sarkar@hpe.com
Hewlett Packard Enterprise
San Jose, USA

## ABSTRACT

The increasing global emphasis on sustainability and reducing carbon emissions is pushing governments and corporations to rethink their approach to data center design and operation. Given their high energy consumption and exponentially large computational workloads, data centers are prime candidates for optimizing power consumption, especially in areas such as cooling and IT energy usage. A significant challenge in this pursuit is the lack of a configurable and scalable thermal data center model that offers an end-to-end pipeline. Data centers consist of multiple IT components whose geometric configuration and heat dissipation make thermal modeling difficult. This paper presents PyDCM, a customizable Data Center Model implemented in Python, that allows users to create unique configurations of IT equipment with custom server specifications and geometric arrangements of IT cabinets. The use of vectorized thermal calculations makes PyDCM orders of magnitude faster (30 times) than current Energy Plus modeling implementations and scales sublinearly with the number of CPUs. Also, PyDCM enables the use of Deep Reinforcement Learning via the Gymnasium wrapper to optimize data center cooling and offers a user-friendly platform for testing various data center design prototypes.

*Equal contribution
†Corresponding author: soumyendu.sarkar@hpe.com

## 1 INTRODUCTION

With the growing demand for cloud computing and data storage, data centers (DCs) are being built at an unprecedented rate [15]. Despite their efficiency in processing data for enterprises, DCs have an enormous energy requirement and carbon footprint. In fact, a moderately sized DC can consume up to a hundred times more energy compared to office spaces of a similar size [14], making them an ideal target for improving energy efficiency and reducing carbon footprint.

The accelerated design cycles associated with these DCs necessitate an approach that can quickly approximate several key performance indicators (KPIs) based on certain parameter choices. Moreover, with the advances in machine learning (ML) and reinforcement learning (RL) in several areas such as healthcare, sustainable energy, and trustworthy AI, the use of these technologies to design optimal control strategies is also an area of active research [3, 11, 12, 16–26].

DC performance relies on several important factors: the geometric configuration of the IT cabinets, the workload, the nature of containment, the airflow through individual servers, the temperature set point and fan speed of the cooling room air conditioning units (CRACs), the design of the evaporator and chiller components, and the cooling tower, among others.

A model that helps incorporate these design options and allows researchers to design and apply RL-based real-time control optimization to DCs in a fast and efficient manner would advance the state of research on efficient DC design, management, and control.

Researchers have been relying on tools such as EnergyPlus [2] or Modelica [10] based models for design and control, which, in combination with interface software such as Sinergym [7] and PyFMU [5], allow for testing of RL-based controllers. While these approaches allow for RL control, there is considerable communication overhead between the modeling platforms and the RL controller, with the latter being commonly designed in the Python programming language. Hence, researchers tend to develop such models in Python for ease of use. One such modeling approach, CityLearn [30], has been very popular among smart grid researchers.

This paper introduces PyDCM, a modular, scalable, and highly customizable DC model suitable for testing a broad range of DC design parameters. PyDCM offers a unique Python-based platform that can be easily adapted for fast prototyping and control optimization by the energy community. Moreover, it serves as a powerful tool for the ML community to test Deep RL initiatives for DC models aimed at sustainability. PyDCM is wrapped with a standard OpenAI Gymnasium interface [28] for rapid testing of DRL algorithms.

## 2 PYDCM ARCHITECTURE

An overview of the PyDCM architecture with the proposed use cases is shown in Figure 1. Model inputs include parameters and settings for various IT equipment and HVAC components, supply and approach temperature results precalculated from a Computational Fluid Dynamics (CFD) simulation [4], and local weather data, grid energy usage, and carbon intensity data. All these parameters and configurations can be chosen by the PyDCM users, enabling them to conduct a DC design analysis to determine hot spots, advanced AI-enabled optimal load allocation, and energy management.

PyDCM integrates different aspects of the DC in a modular, hierarchical, object-oriented design within IT and HVAC models. The IT model, based on the implementations in [27], encompasses the IT room consisting of cabinets filled with servers. Each cabinet is designed as a distinct model, granting users full control over the power and thermal configurations of the IT room. The IT model calculates the power curves for both CPU and fans for every CPU, which are then used to determine the total power consumption of the IT cabinet. This consumption, based on each CPU's utilization and the cabinet inlet temperature, is then used to estimate the overall DC power consumption. The parameters for these curves are also configurable. Additionally, users can customize DC layouts in terms of IT equipment placement geometry and the HVAC system's air containment, including options for cold air containment, hot air containment, and open (no) air containment.

The HVAC model [1] includes a CRAC fan unit to circulate air in the IT room, a chiller for extracting heat from the IT cabinets, a pump for transferring heat to the cooling tower, and a cooling tower for heat dissipation. Each component is modeled separately, allowing users to specify custom parameters for the HVAC system. The HVAC model then estimates the power consumption of each component based on its power and thermal characteristics.

Next, we provide the detailed models implemented in the data center for the software:

*Data Center IT Model:* Let $\tilde{B}_t$ be the DC workload at time instant $t$. The spatial temperature gradient, $\Delta T_{supply}$, given the DC configuration, is obtained from Computational Fluid Dynamics (CFD). For a given rack, the inlet temperature $T_{inlet,i}$ at $CPU_i$ is computed as:

$$T_{inlet,i,t} = \Delta T_{supply,i} + T_{CRACsupply,t} \quad (1)$$

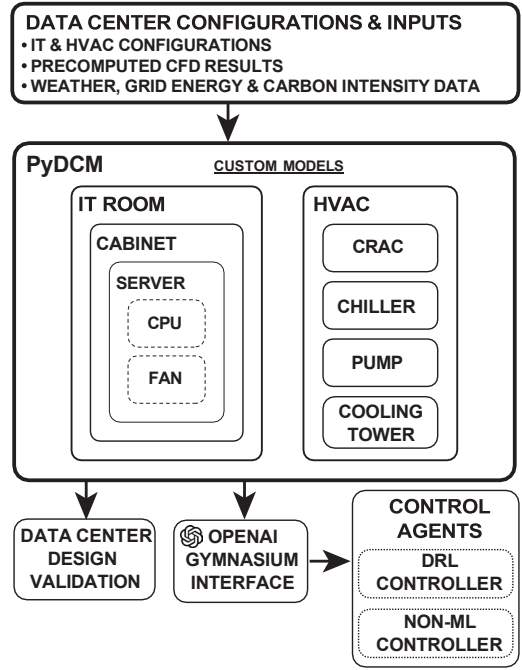

Figure 1: Overview of the PyDCM architecture.

where $T_{CRACsupply,t}$ is the CRAC unit supply air temperature. This value is chosen by an RL agent. Next, the CPU $j$ power curve $f_{cpu,j}(inlet\_temp, cpu\_load)$ and IT Fan power curve $f_{itfan,j}(inlet\_temp, cpu\_load)$ are implemented as linear equations based on [27]. Given a server inlet temperature of $T_{inlet,i,t}$ at IT Cabinet $i$ and a workload of $\tilde{B}_t$ performed by all the $N_i$ CPUs in $i$, the total IT Cabinet $i$ power consumption ($P_{rack,i,t}$), and subsequently the total DC IT Power Consumption ($P_{datacenter,t}$) across all K cabiinets from $i = 1$ to $K$ can be calculated as follows:

$$P_{CPU,i,t} = \sum_{j=1}^{N_i} f_{cpu,j}(T_{inlet,i,t}, \tilde{B}_t) \qquad P_{IT\ Fan,i,t} = \sum_{j=1}^{N_i} f_{itfan,j}(T_{inlet,i,t}, \tilde{B}_t)$$

$$P_{rack,i,t} = P_{CPU,i,t} + P_{IT\ Fan,i,t} \qquad P_{datacenter,t} = \sum_i P_{rack,i,t}$$

The framework also provides an extensive set of models for the HVAC system [1] ($HVAC$) which can be subclassed for designing custom models by the user. The parameters may be set from existing parameter estimation methods to simulate models of actual HVAC components.

*HVAC Cooling Model:* Based on the DC IT Load $P_{datacenter,t}$, the IT fan airflow rate, $V_{sfan}$, air thermal capacity $C_{air}$, and air density, $\rho_{air}$, the rack outlet temperature $T_{outlet,i,t}$ for cabinet $i$ is estimated from [27] using:

$$T_{outlet,i,t} = T_{inlet,i,t} + \frac{P_{rack,k,t}}{C_{air} * \rho_{air} * V_{sfan}} \quad (2)$$

In conjunction with the return temperature gradient information $\Delta T_{return}$ estimated from CFDs, the final CRAC return temperature is obtained as:

$$T_{CRACreturn,t} = avg(\Delta T_{return,i} + T_{outlet,i,t}) \quad (3)$$

We assume a fixed-speed CRAC Fan unit for circulating air through the IT Room. Hence, the total HVAC cooling load for a given CRAC setpoint $T_{CRACsupply,t}$, return temperature $T_{CRACreturn,t}$ and the mass flow rate $m_{crac,fan}$ is calculated as:

$$P_{cool,t} = m_{crac,fan} * C_{air} * (T_{CRACreturn,t} - T_{CRACsupply,t}) \quad (4)$$

To perform $P_{cool,t}$ the amount of cooling, the net chiller load for a chiller with Coefficient of Performance ($COP$) may be estimated as:

$$P_{chiller,t} = P_{cool,t} \left(1 + \frac{1}{COP}\right) \quad (5)$$

This cooling load is serviced by the cooling tower. Assuming a cooling tower delta as a function of ambient temperature $f_{ct\_delta}(T_{ambient,drybulb})$ [1], the required cooling tower air flow rate is calculated as:

$$V_{ct,air,t} = \frac{P_{chiller,t}}{C_{air} * \rho_{air} * f_{ct\_delta}(T_{ambient,drybulb})} \quad (6)$$

Finally, the Cooling Tower Load at a flow rate of $V_{ct,air,t}$ is calculated with respect to a reference air flow rate $V_{ct,air,REF}$ and power consumption $P_{ct,REF}$ from the configuration object:

$$P_{HVAC,cooling,t} = P_{ct,REF} * \left(\frac{V_{ct,air,t}}{V_{ct,air,REF}}\right)^3 \quad (7)$$

The goal of the DC HVAC RL agent is to minimize the total cooling energy and hence the carbon footprint by controlling the $A_{dc,t} = T_{CRACsupply,t}$ given the current CPU workload ($\tilde{B}_t$), weather condition, grid CI ($CI_t$), UPS Battery SoC ($BatSoC$) and other related temporal and spatial information as outlined in the equations above.

## 3 THE LIMITATIONS OF CURRENT IMPLEMENTATIONS

EnergyPlus is recognized for its capabilities in conducting detailed thermal and energy simulations. However, in dynamic contexts such as data center operations, it presents several challenges, underscoring the need for more user-friendly alternatives like PyDCM.

Complexity and Overhead: Configuring and tuning the EnergyPlus input data file (IDF) requires a deep technical understanding of its detailed structure and components. For people unfamiliar with the EnergyPlus environment, including deep reinforcement learning researchers and data center designers, this intricate landscape can be overwhelming. This makes rapid prototyping and understanding the existing models difficult, while hindering the creation of new ones from scratch.

Moreover, to achieve real-time control or to bridge EnergyPlus with modern programming environments, it often becomes necessary to integrate third-party tools like BCVTB [29] or MLE+ [13]. While these tools enable the desired communication, they introduce added layers of complexity and significantly increase communication overhead. This results in slower response times, making real-time simulations and controls more challenging to achieve.

Modification and Processing Time: In the field of dynamic simulations, particularly those guided by RL, the ability to quickly modify and reset simulations is crucial. RL's iterative nature requires frequent environmental adjustments, which may involve altering certain parameters such as simulation intervals, system responses, or input conditions. EnergyPlus, despite its extensive modeling capabilities, has limitations in this regard. Creating a new simulation or even restarting an existing one can be a time-consuming process for EnergyPlus. This is due to the intricacies of the compilations of IDF configurations and the underlying calculations [6].

This lengthy setup and run time can be especially limiting in RL contexts. For example, if an RL agent is trained to optimize cooling strategies for a data center, it will have to interact with the simulated environment thousands, if not millions, of times. Each iteration or adjustment, such as changing the simulated time interval, necessitates a restart of the environment. These time overheads, while they may seem trivial in isolated runs, add up significantly in iterative RL processes, so a more agile, responsive, and efficient platform is needed for data center simulations.

Granularity in Workload Modeling: Current implementations [7, 11, 25, 27] of a data center model with EnergyPlus do not allow the user to specify the individual CPU utilization in an IT Cabinet or specific arrangement of the IT cabinets. Instead, they provide a model that typically permits only one type of heat-generating unit, akin to modeling a single type of "CPU" or server, for the entire room. Consequently, if users aim to represent the intricacies of various server workloads, their options are limited. Additionally, these implementations constrain customizability by mandating a uniform workload across all servers. That is, users define a single CPU utilization value, which is then uniformly applied to all the servers in the data center to estimate IT energy consumption and heat generation.

## 4 PYDCM CONFIGURABILITY AND EXTENSIBILITY

While current implementations of data center models in EnergyPlus present challenges in terms of design flexibility, with constraints such as uniform workload percentages for servers and a single object to describe all servers in a data center zone, PyDCM excels in granular control. In PyDCM, users can specify different characteristics for each CPU in a cabinet (e.g., idle power consumption, full-load power consumption and specific power curves). Users can also set specific utilization percentages for each server and parameters for IT cabinet and HVAC models, thereby laying the foundation for detailed and customized data center simulations. For intricate customizations, we offer a Python environment with user-friendly JSON files. These files are straightforward to interpret and can be effortlessly edited and processed through configuration readers in PyDCM. Moreover, by leveraging an object-oriented modeling (OOM) approach, PyDCM empowers advanced users to subclass and craft their own data center components.

PyDCM significantly reduces communication overhead with other machine learning applications in Python compared to EnergyPlus, as it avoids the need for a cross-platform interface. Additionally, PyDCM utilizes vectorized computations for data center dynamics and conducts in-place operations, optimizing both computational speed and memory usage. The core models in the `datacenter` and `hvac_models` classes employ this vectorization to provide maximum performance enhancement, especially as the number of servers and cabinets increases.

Furthermore, relative to existing implementations, PyDCM offers an efficient simulation `reset`—a process in RL executed at the beginning of each training episode—and a `step` method optimized for speedy execution. Conversely, EnergyPlus exhibits notable reset

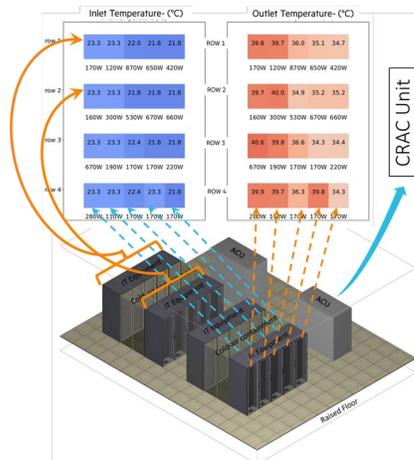

**Figure 2: Temperature Distribution with Cold Air Containment**

| Parameter | Description | Example Value |
|---|---|---|
| NUM_ROWS | # of rows in the data center | 5 |
| NUM_RACKS_PER_ROW | # of racks per row | 10 |
| CPUS_PER_RACK | # of CPUs per rack | 40 |
| RACK_SUPPLY_APPROACH_TEMP_LIST | Supply temperatures for each rack | [22, 22.5, ...] |
| C_AIR | Air properties | 1006 |
| CHILLER_COP | Chiller's coefficient of performance | 6.0 |
| IT_FAN_AIRFLOW_RATIO_LB | LB Fan airflow ratios for IT equipment | [0.0 0.6] |
| IT_FAN_AIRFLOW_RATIO_UB | UB Fan airflow ratios for IT equipment | [0.7 1.3] |

**Table 1: PyDCM Configuration Parameters and Descriptions**

latencies, primarily due to the generation of auxiliary files needed for the simulation.

### 4.1 Getting Started with PyDCM

We show a basic data center simulation using PyDCM by modifying the JSON file `utils/dc_config.json`. Table 1 outlines some parameters included in the JSON configuration file:

With these configurations in place, initializing and running a PyDCM simulation is as simple as: 1) Modify the Configuration. 2) Initialize Environment: Use the configuration to create the PyDCM environment. 3) Run Simulation as an OpenAI Gymnasium environment. Users can effortlessly set up a custom data center simulation with PyDCM, taking advantage of the platform's customizability and efficiency. For a more detailed tutorial, please refer to the provided notebook[1]. The notebook points to the current PyDCM repository.

## 5 COMPARATIVE ANALYSIS WITH CURRENT DATA CENTER MODELS

### 5.1 Comparison with RL applications

We benchmarked PyDCM against the current data center implementation in EnergyPlus [11, 27] by focusing on three RL methods: "init", "reset", and "step". The cumulative simulation times, combining "reset" and "step", were also analyzed for different episode lengths: 7 and 30 days. The simulation time step was 15 minutes.

---

[1]Link to **Colab Notebook for PyDCM**

| Method | EnergyPlus | PyDCM | Reduction (%) |
|---|---|---|---|
| init | 1.05s ± 23.6ms | 1.57ms ± 60.4μs | 99.85 |
| reset | 2.67s ± 23.8ms | 0.03ms ± 0.25μs | 99.99 |
| step | 0.46ms ± 98.38μs | 0.13ms ± 15.84μs | 71.33 |

**Table 2: Comparison of method timings between EnergyPlus and PyDCM. Mean ± std. dev. of 10 simulations.**

| Episode | EnergyPlus | PyDCM | Reduction (%) |
|---|---|---|---|
| 30 days | 3.33s ± 91.20ms | 0.34s ± 42.20ms | 89.79 |
| 7 days | 2.64s ± 34.39ms | 0.09s ± 1.86ms | 96.77 |

**Table 3: Total simulation time comparison for different RL episode lengths. Mean ± std. dev. of 10 simulations.**

| Metric | EnergyPlus | PyDCM | Reduction (%) |
|---|---|---|---|
| Wait. Time | 1.48s ± 0.22s | 0.27s ± 0.48ms | 81.55 |
| Sample Time | 9.28s ± 0.51s | 3.95s ± 16.20ms | 57.34 |

**Table 4: Comparison of Performance Metrics for RL Environments. Mean ± std. dev. of 10 simulations.**

All tests were carried out in a data center with two zones, as demonstrated in [11, 27]. In our evaluation, detailed in Table 2, PyDCM showcased significant improvement across the "init", "reset", and "step" RL methods. PyDCM's acceleration can be attributed to its utilization of vectorized and in-place computations for data center dynamics, which optimizes both memory and compute time. The total simulation times for different episode lengths are summarized in Table 3. The individual improvements in the step and reset methods lead to cumulative improvements.

### 5.2 Scalability

To assess scalability, we conducted a series of simulations (Figure 3), progressively increasing the number of simulated CPUs and tracking the total simulation time.

Performance Enhancement with PyDCM: PyDCM significantly outperforms the existing EnergyPlus implementation for data center simulations. Specifically, PyDCM can operate at speeds more than 40 times faster than EnergyPlus. When examining hyper-scale data centers—characterized by more than 10,000 CPUs (denoted with a vertical line as "Hyper-Scale DC" [8])—PyDCM is able to reduce the simulation times by a factor of 16.

Underlying Assumptions in EnergyPlus Calculations: An interesting pattern emerged when observing the consistent simulation time across different CPU counts in EnergyPlus. This behavior suggests that the EnergyPlus model might be built on certain assumptions. It could calculate the energy and thermal properties of a single CPU and then linearly scale (multiply) this base value by the total number of CPUs. While such a method can streamline calculations, it limits customization.

### 5.3 Resource Utilization Analysis

In terms of system resources, while optimizing control using RLLib [9], EnergyPlus uses 18.20$GB$ of RAM, while PyDCM uses a slightly lower 16.84$GB$. Moreover, PyDCM's CPU utilization is more efficient, registering at 18.21%, as opposed to EnergyPlus's 20.64%.

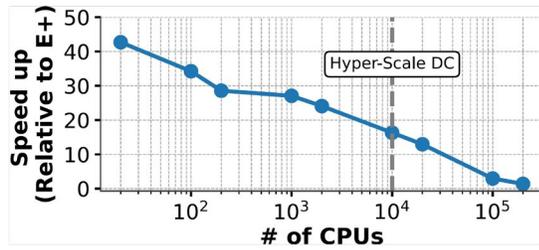

**Figure 3: Simulation speed up relative to the current implementation of EnergyPlus (E+). * Hyper-scale data center consists on more than 10,000 CPUs. [8]**

These experiments were conducted on a server equipped with a 48-core Intel Xeon 6248 CPU.

### 5.4 Reinforcement Learning Metrics Analysis

We compared two pivotal RL metrics (provided by RLLib) between EnergyPlus and PyDCM: Sample Time (time for workers to accumulate samples for one training iteration) and waiting time (delay when waiting for the environment during polling). Table 4 indicates that PyDCM achieves an 81.55% reduction in waiting time and a 57.34% decrease in sampling time compared to EnergyPlus implementations.

We also used PyDCM to develop a DRL agent to control the HVAC setpoint, achieving an 8% reduction in energy consumption and carbon emissions in data centers.

## 6 CONCLUSION AND FUTURE WORK

PyDCM with its Pythonic vectorized implementation provides a fast and effective tool for designing highly customizable data center prototypes with an easy integration with machine learning framework. This enables research on optimal control using deep reinforcement learning. We plan to incorporate IoT sensors for parameter refinements for digital twins.